%% file: main.tex
\documentclass[a4paper,10pt]{svproc}
\usepackage{graphicx,psfrag,booktabs,amsmath,amsfonts,verbatim}
\usepackage{cases}
\usepackage{url}

\usepackage{algorithm}
\usepackage{pgfplots}
\usepackage{tikz}
\usepackage{tikzscale}  
\usetikzlibrary{arrows.meta}
\pgfplotsset{compat=1.17}
\usepackage{support-caption}
\usepackage{subcaption}
\usepackage{caption}
\usepackage{algpseudocode}


\usepackage[numbers]{natbib}
\usepackage{bbding}
\usepackage{pifont}
\usepackage[title,toc]{appendix}
\usepackage{wrapfig}

\let\llncssubparagraph\subparagraph
\let\subparagraph\paragraph
\usepackage[compact]{titlesec}
\let\subparagraph\llncssubparagraph
\titlespacing*{\section}{0pt}{0.75\baselineskip}{\baselineskip}
\titlespacing*{\subsection}{0pt}{0.75\baselineskip}{\baselineskip}
\titlespacing*{\subsubsection}{0pt}{0.75\baselineskip}{\baselineskip}

\renewcommand\bibpreamble{\vspace{-0.8\baselineskip}} 
\renewcommand{\refname}{References}

\def\checkmark{\tikz\fill[scale=0.4](0,.35) -- (.25,0) -- (1,.7) -- (.25,.15) -- cycle;}

\usepackage{tabstackengine}
\stackMath

\author{Taylor Howell, Kevin Tracy, Simon Le Cleac'h, and Zachary Manchester}

\begin{document}
\mainmatter
\title{CALIPSO: A Differentiable Solver \\ for Trajectory Optimization with \\ Conic and Complementarity Constraints}
\titlerunning{CALIPSO}
\author{Taylor A. Howell\inst{1} \and Kevin Tracy\inst{2} \and Simon Le Cleac'h\inst{1} \and Zachary Manchester\inst{2}}
\authorrunning{Howell et al.}

\tocauthor{Taylor Howell, Kevin Tracy, Simon Le Cleac'h, and Zachary Manchester}
\institute{Stanford University, Stanford CA 94305, USA,\\
\email{thowell@stanford.edu},\\ WWW home page:
\texttt{http://roboticexplorationlab.org}
\and
Carnegie Mellon University, Pittsburgh PA 15213, USA}

\maketitle

\begin{abstract}
We present a new solver for non-convex trajectory optimization problems that is specialized for robotics applications. CALIPSO, or the Conic Augmented Lagrangian Interior-Point SOlver, combines several strategies for constrained numerical optimization to natively handle second-order cones and complementarity constraints. It reliably solves challenging motion-planning problems that include contact-implicit formulations of impacts and Coulomb friction and state-triggered constraints where general-purpose non-convex solvers like SNOPT and Ipopt fail to converge. Additionally, CALIPSO supports efficient differentiation of solutions with respect to problem data, enabling bi-level optimization applications like auto-tuning of feedback policies. Reliable convergence of the solver is demonstrated on a range of problems from manipulation, locomotion, and aerospace domains. An open-source implementation of this solver is available.

\end{abstract}


\section{Introduction}

Trajectory optimization is a powerful tool for offline generation of complex behaviors for dynamic systems, as well as online as a planner or feedback controller within model predictive control frameworks.  The use of constraints greatly enhances the ability of a designer to generate desirable solutions, enforce safe behaviors, and model physical phenomena. Unfortunately, many constraint types that have important and direct applications to robotics are poorly handled by existing general-purpose non-convex solvers \cite{nocedal2006numerical} or differential dynamic programming (DDP) algorithms \cite{jacobson1970differential}. 

Second-order cones \cite{boyd2004convex}, which commonly appear as friction-cone \cite{moreau2011unilateral} or thrust-limit \cite{blackmore2010minimum} constraints, present difficulties for these solvers due to their nondifferentiability at commonly occurring states, like when the friction or thrust forces are zero. Common reformulations of these constraints for solvers like SNOPT \cite{gill2005snopt} and Ipopt \cite{wachter2006implementation} are typically non-convex and fail to work well in practice \cite{vanderbei1998using}.

Contact dynamics, including impact and friction, are naturally modeled with complementarity constraints \cite{scheel2000mathematical}. This formulation constrains contact forces to only take on non-zero values when the distance between objects is zero. State-triggered constraints \cite{szmuk2019real}, in which constraints switch on or off in different regions of the state space, can similarly be modeled with complementarity constraints. However, these constraints violate the linear independence constraint qualification (LICQ), a fundamental assumption in the convergence theory of standard second-order solvers \cite{nocedal2006numerical, biegler2010nonlinear}.

In this work, we present a new solver for trajectory optimization, CALIPSO: Conic Augmented Lagrangian Interior-Point SOlver. The development of this solver is motivated by challenging non-convex motion-planning problems that require second-order cone and complementarity constraints; in particular, contact-implicit trajectory optimization \cite{posa2014direct, manchester2020variational} for locomotion and manipulation. The solver combines a number of ideas and algorithms from constrained numerical optimization to solve these difficult problems reliably. 

Second-order cones are handled using an interior-point method \cite{vandenberghe2010cvxopt} that exploits the convexity of these constraints for strict enforcement without the need for linear approximations. All equality constraints are handled using an augmented Lagrangian method \cite{bertsekas2014constrained}, which does not require the constraints to satisfy LICQ and is robust to the degeneracies that can arise from complementarity constraints \cite{izmailov2012global}. Additionally, both the interior-point and augmented-Lagrangian methods are formulated in a primal-dual fashion to enhance the numerical robustness and performance of the solver. The computation of Newton steps on the combined primal-dual augmented-Lagrangian and interior-point KKT conditions are reformulated as a symmetric linear system to enable the use of fast, direct linear algebra methods \cite{davis2005algorithm}. Finally, the implicit-function theorem \cite{dini1907lezioni,amos2017optnet} is utilized to efficiently differentiate through solutions, enabling bi-level optimization applications such as auto-tuning of model predictive control policies.

Our specific contributions are:
\begin{itemize} 
	\item A differentiable trajectory optimization solver with native support for second-order cones and reliable handling of complementarity constraints
	\item A novel, combined, interior-point and augmented-Lagrangian algorithm for non-convex optimization
	\item A symmetric reformulation of the combined method's KKT system for fast symmetric linear-system solvers
	\item A collection of benchmark robot motion-planning problems that contain second-order cone and complementarity constraints
	\item An open-source implementation of the solver written in Julia
\end{itemize}

In the remainder of this paper, we first provide an overview of related work on trajectory optimization methods in Section \ref{related_work}. Then, we provide the necessary background on complementarity constraints, LICQ,  augmented Lagrangian methods, and interior-point methods in Section \ref{background}. Next, we present CALIPSO and its key algorithms and routines in Section \ref{calipso}. We then demonstrate CALIPSO on a collection of robot motion-planning examples in Section \ref{results}. Finally, we conclude with a discussion of limitations and directions for future work in Section \ref{conclusion}.

\section{Related work} \label{related_work}
Trajectory optimization problems are solved with methods that are classically categorized as either indirect or direct \cite{betts1998survey}. Indirect methods include shooting methods, DDP \cite{jacobson1970differential}, and iterative LQR (iLQR) \cite{li2004iterative}. These methods exploit the temporal structure of the problem and utilize a Riccati backward pass to compute updates for control variables followed by forward simulation rollouts to update the states. Classically, apart from the dynamics, these methods do not include support for constraints. In recent years, various approaches have extended these methods to handle constraints \cite{tassa2014control,howell2019altro, howell2022trajectory,mastalli2020crocoddyl,singh2022optimizing,jallet2022constrained,jackson2021altro}. However, reliable constraint handling and solution accuracy for these methods is still challenging in many scenarios.

Direct methods, in contrast, directly transcribe the trajectory optimization problem as a constrained non-convex problems, with both states and controls as decision variables \cite{stryk1993numerical}. The transcription is provided to a general-purpose solver such as SNOPT or Ipopt. This approach is generally reliable and enables robust constraint handling. However, these solvers do not exploit the temporal structure of the trajectory optimization problem like indirect methods, and historically have been thought to converge to solutions more slowly as a result. To address this limitation, direct solvers tailored for the underlying trajectory optimization problem structure have been proposed \cite{wang2009fast} and are available as commercial tools \cite{zanelli2020forces}. 

While providing reliable constraint handling for equality and inequality constraints in many scenarios, direct methods lack support for second-order cones \cite{boyd2004convex} and exhibit difficulties handling complementarity constraints \cite{scheel2000mathematical}. The exact handling of second-order cones as inequality constraints with these solvers results in poor practical performance because the nondifferentiable point of the cone is frequently visited in robotics applications (e.g., an object resting on a surface with zero friction force). Classic reformulations are non-convex and similarly exhibit poor and unreliable convergence \cite{vanderbei1998using}. Recently, sequential convexification approaches have been developed that are able to handle second-order cones by iteratively solving convex approximations of the original problem using general-purpose cone solvers \cite{szmuk2020successive,bonalli2019gusto}. 

The complementarity constraints that can arise in robotics problems are generally difficult for solvers to handle because they are non-convex and violate LICQ. To overcome this, a number of problem reformulations and constraint relaxation techniques have been proposed and explored for interior-point \cite{biegler2010nonlinear,raghunathan2003mathematical} and augmented Lagrangian \cite{izmailov2012global} methods, but these approaches largely remain \textit{ad hoc} and are unavailable in state-of-the-art solvers for trajectory optimization, which we compare in Table \ref{trajopt_solvers}.

\begin{table}[t]
	\centering
	\caption[Comparison of optimizers]{Comparison of general-purpose and trajectory optimization solvers.}
	\scriptsize
	\begin{tabular}{c c c c c c}
		\toprule
		\textbf{Solver} &
		\textbf{Method} &
		\textbf{Accuracy} &
		\textbf{Second-Order} &
		\textbf{Complementarity} &
		\textbf{Differentiable} \\
		\toprule
		Ipopt \cite{wachter2006implementation} & direct & high & \ding{53} & \ding{53} & \ding{53} \\
		SNOPT \cite{gill2005snopt} & direct & high & \ding{53} & \ding{53} & \ding{53} \\
		CVX \cite{agrawal2019differentiable} & direct & high & \checkmark & \ding{53} & \checkmark \\
		\toprule
		ALTRO \cite{howell2019altro} & indirect & medium & \checkmark & \ding{53} & \ding{53} \\
		Trajax \cite{trajax} & indirect & medium & \ding{53} & \ding{53}  & \checkmark \\
		GuSTO \cite{bonalli2019gusto} & direct & high & \checkmark & \ding{53} & \ding{53} \\
		FORCES \cite{zanelli2020forces} & direct & high & \ding{53} & \ding{53} & \ding{53} \\
		Drake \cite{drake} & direct & high & \ding{53} & \ding{53} & \ding{53} \\
		\toprule
		CALIPSO & direct & high & \checkmark & \checkmark & \checkmark \\
		\toprule
	\end{tabular}
	\label{trajopt_solvers}
\end{table}

\section{Background} \label{background}
In this section, we first provide a brief overview of trajectory optimization, followed by background on complementarity constraints, LICQ, augmented Lagrangian methods, and interior-point methods for constrained optimization. Finally, we compare approaches for differentiating through a solver. For helpful background and notation details, see \cite{boyd2004convex}.

\subsection{Trajectory optimization}
Direct methods for trajectory optimization transcribe problems into standard instances:
\begin{equation}
	\begin{array}{ll}
	\underset{x}{\mbox{minimize }}  & c(x) \\
	\mbox{subject to } & g(x) = 0, \\
					  & h(x) \in \mathcal{K},
	\end{array} \label{general_problem}
\end{equation}
with decision variables $x \in \mathbf{R}^n$, objective $c : \mathbf{R}^n \rightarrow \mathbf{R}$, equality constraints $g : \mathbf{R}^n \rightarrow \mathbf{R}^m$, and inequality constraints $h : \mathbf{R}^n \rightarrow \mathbf{R}^p$ in cone $\mathcal{K}$. The functions are assumed to be smooth and twice differentiable, and the cone is the Cartesian product of convex cones (e.g., standard positive-orthant inequalities and second-order cones) \cite{boyd2004convex}. 

Trajectory optimization problems:
\begin{equation}
	\begin{array}{ll}
		\underset{X_{1:T}, U_{1:T-1}}{\mbox{minimize }} & C_T(X_T) + \sum \limits_{t = 1}^{T-1} C_t(X_t, U_t)\\
		\mbox{subject to } & F_t(X_t, U_t) = X_{t+1},\phantom{\mathcal{K},} \quad t = 1,\dots,T-1,\\
		& E_t(X_t, U_t) = 0,\phantom{\,_{t+1}\mathcal{K}_t}\quad t = 1, \dots, T,\\
		& H_t(X_t, U_t) \in \mathcal{K}_t,\phantom{\,x_{t+1}}\quad t = 1, \dots, T, \label{trajopt}\\
	\end{array}
\end{equation}
are instances with special temporal structure for a dynamical system with state $X_t \in \mathbf{R}^{n_t}$, control inputs $U_t \in \mathbf{R}^{m_t}$, time index $t$, discrete-time dynamics $F_t : \mathbf{R}^{n_t} \times \mathbf{R}^{m_t} \rightarrow \mathbf{R}^{n_{t+1}}$, and stage-wise objective $C_t: \mathbf{R}^{n_t} \times \mathbf{R}^{m_t} \rightarrow \mathbf{R}$,  equality constraints $E_t : \mathbf{R}^{n_t} \times \mathbf{R}^{m_t} \rightarrow \mathbf{R}^{e_t}$, and cone constraints $H_t : \mathbf{R}^{n_t} \times \mathbf{R}^{m_t} \rightarrow \mathbf{R}^{h_t}$, 
over a planning horizon $T$.

\subsection{Complementarity constraints}
Contact-implicit trajectory optimization \cite{posa2014direct} optimizes trajectories for systems that make and break contact with their environments and represents dynamics using complementarity constraints. For example, optimizing motion over a single time step for an actuated particle in a single dimension, resting on a surface, modeled with impact such that it cannot pass through the floor:
\begin{equation}
	\begin{array}{ll}
		\underset{z, u, \gamma}{\mbox{minimize }} & \frac{1}{2}(z - z_{\mbox{g}})^2 + \frac{1}{2}u^2 \\
		\mbox{subject to } & m \big(z / h + g h \big) = \gamma + u, \\
		& z \cdot \gamma = 0, \\
		& z, \gamma \geq 0,
	\end{array} \label{calipso_ci_trajopt}
\end{equation}
with position $z\in \mathbf{R}$, control input $u \in \mathbf{R}$, contact impulse $\gamma \in \mathbf{R}$, mass $m \in \mathbf{R}_{++}$, gravity $g \in \mathbf{R}_{+}$, time step $h \in \mathbf{R}_+$, and goal $z_{\mbox{g}} \in \mathbf{R}$. These constraints are derived from a constrained discrete Lagrangian \cite{manchester2020variational}. The set of constraints on $z$ and $\gamma$ are collectively referred to as a complimentary constraint, and are sometimes abbreviated $z \perp \gamma$. This formulation does not require pre-specified contact-mode sequences or hybrid dynamics since the solver is able to optimize physically correct contact dynamics at each time step.

\subsection{Linear independence constraint qualification}
General-purpose second-order solvers that rely on Newton's method to compute search directions (e.g., SNOPT and Ipopt) assume that the constraints provided by the user satisfy the LICQ in the neighborhood of solutions. Certain classes of constraints, including complementarity conditions that naturally arise in contact dynamics, often do not satisfy this assumption. 
We demonstrate how this assumption is violated with a simple contact-implicit trajectory optimization problem \eqref{calipso_ci_trajopt}. The Lagrangian for the problem is: 
\begin{equation}
	L(z, u, \gamma, a, b, c, d) = \frac{1}{2}(z - z_{\mbox{g}})^2 + \frac{1}{2}u^2 + a \big( m (z / h + g h ) - \gamma - u \big) + b z \gamma - c z - d \gamma,
\end{equation}
where $a, b, c, d \in \mathbf{R}$ are the Lagrange multipliers, or ``dual variables,'' associated with constraints. The KKT system is:

\begin{equation} 
	\setstackgap{L}{1.1\baselineskip}
	\fixTABwidth{T}
	\bracketMatrixstack{
		1 & b & 0 & m / h & \gamma & -1 & 0 \\
		b & 0 & 0 & 1 & z & 0 & -1 \\
		0 & 0 & 1 & -1 & 0 & 0 & 0 \\
		m / h & -1 & -1 & 0 & 0 & 0 & 0 \\
		\gamma & z & 0 & 0 & 0 & 0 & 0 \\
		c & 0 & 0 & 0 & 0 & z & 0 \\ 
		0 & d & 0 & 0 & 0 & 0 & \gamma
	}
	\bracketMatrixstack{
		\Delta z\\
		\Delta \gamma \\
		\Delta u \\
		\Delta a \\
		\Delta b \\
		\Delta c \\
		\Delta d
	}
	=
	-
	\bracketMatrixstack{
		(z - z_{\text{g}}) + a m / h + b \gamma - c \\ 
		a + b z - d \\
		u - a \\ 
		m \big(z / h + g h \big) - \gamma - u \\
		z \cdot \gamma \\ 
		c \cdot z \\ 
		d \cdot \gamma 
	}, 
	\label{calipso_licq_kkt}
\end{equation}
\begin{equation}
	z, \gamma, c, d, \geq 0,
\end{equation}
with first-order necessary (KKT) conditions (right-hand side) and KKT matrix (left-hand side). The Newton step that a standard second-order solver would take to drive these KKT conditions to zero is computed by solving this system \eqref{calipso_licq_kkt}.

In the scenario where the particle is above the surface (i.e., $z > 0$) the contact impulse must be zero (i.e., $\gamma = 0$). As a result, the fifth and seventh rows of the KKT matrix will be linearly dependent, resulting in non-unique optimal dual variables and violation of LICQ. A similar result occurs when $\gamma > 0$ and $z = 0$. Consequently, the Newton step is not well defined in these cases, causing difficulties for the solver. While a myriad of \textit{ad hoc} heuristics exist to alleviate this problem, we pursue a more rigorous approach in the following sections based on ideas from constrained numerical optimization.

\subsection{Augmented Lagrangian methods}
	We now consider equality constrained problems of the form:
	\begin{equation}
		\begin{array}{ll}
	    \underset{x}{\mbox{minimize }}  & c(x) \\
		\mbox{subject to } & g(x) = 0. \\
		\end{array}
		\label{equality_constrained}
	\end{equation}
	The augmented Lagrangian method transforms this problem \eqref{equality_constrained} into an unconstrained problem by introducing dual variables, $\lambda \in \mathbf{R}^m$, and a quadratic penalty parameterized by $\rho \in \mathbf{R}_+$:
	\begin{equation}
		L_{\mathcal{A}}(x; \lambda, \rho) = c(x) + \lambda^T g(x) + \frac{\rho}{2} g(x)^T g(x), 
		\label{augmented_lagrangian}
	\end{equation}
where we refer to $L_{\mathcal{A}}$ as the \emph{augmented Lagrangian} for the problem \eqref{equality_constrained}.

\subsubsection{Primal method.}
The classic method alternates between minimizing the augmented Lagrangian  \eqref{augmented_lagrangian} and performing outer updates on the dual variables and penalty:
\begin{equation} 
	\lambda \leftarrow \lambda + \rho g(x), \quad \rho \leftarrow \phi (\rho), \label{augmented_lagrangian_update}
\end{equation}
until a solution to the original problem \eqref{equality_constrained} is found \cite{bertsekas2014constrained}. This typically requires ten or fewer outer updates and a simple update, $\phi : \mathbf{R}_+ \rightarrow \mathbf{R}_+$, that scales the penalty by a constant value, works well in practice.
Throughout, subscripts are used to denote derivatives and we drop the variable dependence of the functions for clarity. The KKT system for this method is:
\begin{equation} 
	\Big[c_{xx} + \rho g_x^T g_x + \sum \limits_{i = 1}^m (\lambda^{(i)} + \rho g^{(i)}) g_{xx}^{(i)} \Big] \Delta x = -\Big[c_x + g_x^T (\lambda + \rho g)\Big]. \label{augmented_lagrangian_gradient} 
\end{equation}

Search directions $\Delta x \in \mathbf{R}^n$ are computed by solving the linear system \eqref{augmented_lagrangian_gradient} for fixed values of the dual variables and penalty. Newton's method with a line search is utilized to compute iterates that satisfy the KKT conditions, or residual \eqref{augmented_lagrangian_gradient}, to a desired tolerance.

Importantly, the KKT matrix becomes increasingly ill-conditioned as the penalty is increased in order to achieve better satisfaction of the equality constraints, degrading the quality of the Newton step.

\subsubsection{Primal-dual method.}
    To address the deficiencies of the primal method, a primal-dual method introduces additional dual variables, $y \in \mathbf{R}^m$ and constraints:
	\begin{equation}
	y = \lambda + \rho g(x), \label{dual_constraint}
	\end{equation}
	in order to utilize an alternative KKT system with better numerical properties.
	
	Combining these constraints \eqref{dual_constraint} with the primal KKT system \eqref{augmented_lagrangian_gradient} and performing a simple manipulation yields 
	the \textit{primal-dual} augmented-Lagrangian KKT system:
	\begin{align}
	\setstackgap{L}{1.1\baselineskip}
	\bracketMatrixstack{
    	c_{xx} + \sum \limits_{i = 1}^m y^{(i)} g_{xx}^{(i)} & g_x^T \\ 
    	g_x & -\frac{1}{\rho} I
	}
	\bracketVectorstack{ 
    	\Delta x \\ 
    	\Delta y
	}
	= 
	-\bracketMatrixstack{ 
    	c_x + g_x^T y \\ 
    	g + \frac{1}{\rho}(\lambda - y) 
	}. \label{augmented_lagrangian_primal_dual}
	\end{align}
    In contrast to the primal system \eqref{augmented_lagrangian_gradient}, this system \eqref{augmented_lagrangian_primal_dual} does not become ill-conditioned as the penalty is increased because this term does not appear in the KKT matrix---only its inverse appears (i.e., $-\frac{1}{\rho} I$), which actually enhances the conditioning of the system by performing dual regularization  \cite{kuhlmann2018primal,gill2012primal,argaez2002global}. Further, this method does not require LICQ because the KKT matrix remains full rank even in cases where $g_x$ is rank deficient as a result of the dual regularization \cite{izmailov2012global}.
    
    Additionally, the KKT conditions now contain relaxed constraints (i.e., $g(x) + \frac{1}{\rho}(\lambda - y))$ that are particularly helpful for complementarity formulations since they are only satisfied in the convergence limit as outer updates are performed. In the contact-dynamics setting, this relaxation corresponds to ``soft'' contact models that are iteratively updated to become ``hard'' as the algorithm converges. 
	
	\subsection{Interior-point methods}
    
	To handle problems with cone constraints (i.e., inequalities and second-order cones) that commonly occur in robotics applications, for example torque limits or friction cones, we employ interior-point methods. To illustrate the approach, we now consider problems with inequality constraints:
	\begin{equation}
		\begin{array}{ll}
			\underset{x}{\mbox{minimize }}  & c(x) \\
			\mbox{subject to } & h(x) \ge 0. \\
		\end{array}
		\label{inequality_problem}
	\end{equation}
	An unconstrained problem is formed by introducing a logarithmic barrier, relaxed by a central-path parameter $\kappa \in \mathbf{R}_{+}$:
	\begin{equation}
		L_{\mathcal{B}}(x; \kappa) = c(x) - \kappa \sum \limits_{i = 1}^p \mbox{log}(h^{(i)}(x)),
		\label{barrier_function}
	\end{equation}
	where we refer to $L_{\mathcal{B}}$ as the \textit{barrier Lagrangian} for the problem \eqref{inequality_problem}.

	\subsubsection{Primal method.}
	The classic method alternates between minimizing the barrier Lagrangian \eqref{barrier_function} and outer updates to the central-path parameter until a solution to the original problem \eqref{inequality_problem} is found \cite{boyd2004convex}. An effective strategy for the update is to decrease the parameter by a constant factor. 
	
	The KKT system for this method is:
	\begin{equation}
	\Big[c_{xx} - \kappa \sum \limits_{i = 1}^p (\frac{1}{h^{(i)}} h_{xx}^{(i)} - \frac{1}{(h^{(i)})^2} (h_x^{(i)})^2) \Big] \Delta x = -\Big[c_x - \kappa \sum \limits_{i = 1}^p \frac{1}{h^{(i)}} h_x^{(i)}\Big]. \label{barrier_gradient}
	\end{equation}
	 As the central-path parameter is decreased, the logarithmic barrier becomes a closer approximation to the indicator function, which has an infinite cost if a constraint is violated and is otherwise zero \cite{boyd2004convex}. While simple, this approach suffers from numerical ill-conditioning, similar to the primal augmented Lagrangian method, as the central-path parameter approaches zero, degrading a solver's ability to find accurate solutions to the original problem \eqref{inequality_problem}.
	
	\subsubsection{Primal-dual method.} 
    
	To address the conditioning issues of the primal method, additional dual variables, $z \in \mathbf{R}^p$, and constraints, $z^{(i)} = \kappa /h^{(i)}(x)$ for $i = 1, \dots, p$, are introduced to form a new KKT system:
	\begin{align} 
	    \setstackgap{L}{1.25\baselineskip}
	    \begin{bmatrix} 
	    c_{xx} - \sum \limits_{i = 1}^p z^{(i)} h_{xx}^{(i)}& h_x^T \\ 
	    \mbox{\textbf{diag}}(z) h_x & \mbox{\textbf{diag}}(h)
	    \end{bmatrix}
	    \begin{bmatrix} 
	    \Delta x \\ 
	    \Delta z
	    \end{bmatrix} &= 
	    -
	    \bracketMatrixstack{ 
    	 c_x - h_x^T z \\
    	 z \circ h - \kappa \mathbf{e}
    	 }, \label{interior_point_primal_dual}\\
    	 z, h &\geq 0 \label{interior_point_inequality}.
	\end{align}
	Similar to the primal-dual augmented Lagrangian method, this KKT matrix does not depend on the central-path parameter, resulting in significantly better numerical conditioning than primal methods. The complementarity constraints in the KKT conditions are relaxed (i.e., $z \circ h(x) - \kappa \mathbf{e})$, only being satisfied in the limit as the central-path parameter is decreased to zero. Here, the target $\mathbf{e}$ is a vector of ones and the cone product $\circ$ denotes an element-wise product.
	
	\subsubsection{Second-order cone constraints.}
    
\begin{figure}[t]
    	\centering
        \includegraphics[width=.3\textwidth]{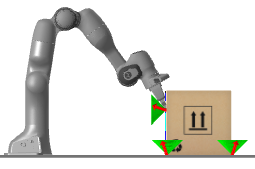}
    	\includegraphics[width=.175\textwidth]{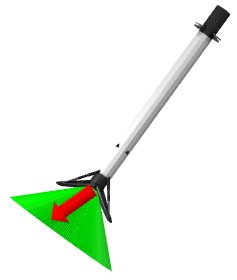}
    	\caption{Robotics applications with second-order cone constraints include Coulomb friction at contact points in manipulation tasks (left) and thrust limits on rockets (right). Cone constraints are shown in green, while force vectors are shown in red.}
    	\label{second_order_cone_robots}
    \end{figure}

	The barrier formulation accommodates second-order cone constraints:
	\begin{equation}
		a \in \mathcal{Q}_l = \{(a^{(1)}, a^{(2:l)}) \in \mathbf{R} \times \mathbf{R}^{l-1}\, | \, \|a^{(2:l)}\|_2 \leq a^{(1)} \}, \label{soc_set}
	\end{equation}
	of dimension $l$, which frequently appear in robotics applications as friction cones and thrust limits (Fig. \ref{second_order_cone_robots}).
    In this setting, the barrier contains squared cone variables:
    \begin{equation} 
        \frac{1}{2} \mbox{log}\Big((a^{(1)})^2 - (a^{(2:l)})^T a^{(2:l)} \Big),
    \end{equation}
    and remains convex; the cone product and target are:
	\begin{align} 
	a \circ b &= (a^T b, \, a^{(1)} b^{(2:l)} + b^{(1)} a^{(2:l)}), \label{soc_product} \\
    \mathbf{e} &= (1, 0_{l-1}). \label{soc_target}
	\end{align}
	The inequality constraints \eqref{interior_point_inequality} are replaced with their second-order cone counterparts \eqref{soc_set} \cite{domahidi2013ecos, vandenberghe2010cvxopt}. 
	
	\subsection{Differentiable optimization} 
    Solvers can be differentiated by unrolling the algorithm and utilizing the chain rule to differentiate through each iterate \cite{domke2012generic}. However, in practice, this approach requires truncating the number of iterates, which can lead to low-accuracy solutions. Additionally, the approach can be plagued by numerical issues that lead to exploding or vanishing gradients. A more efficient approach is to utilize the implicit-function theorem at a solution point in order to compute the sensitivities of the solution \cite{amos2017optnet,agrawal2019differentiating} . 
    
    An implicit function, $R : \mathbf{R}^a \times \mathbf{R}^b \rightarrow \mathbf{R}^a$, is defined as:
    \begin{equation}
    R(w^*; \theta) = 0, \label{implicit_function}
    \end{equation}
    for solutions $w^* \in \mathbf{R}^a$ and problem data $\theta \in \mathbf{R}^b$. At an equilibrium point, $w^*(\theta)$, the sensitivities of the solution with respect to the problem data, i.e., $\partial w / \partial \theta$, can be computed under certain conditions \cite{dini1907lezioni} as:
    \begin{equation}
        \frac{\partial w}{\partial \theta} = -\Big(\frac{\partial R}{\partial w}\Big)^{-1} \frac{\partial R}{\partial \theta}. \label{solution_sensitivity}
    \end{equation}
    In the case that $\partial R / \partial w$ is not full rank, an approximate solution, e.g., least-squares, can be computed. A similar approach for differentiating through DDP-based solvers using a Riccati approach has been proposed \cite{jin2020pontryagin}. In the context of trajectory optimization, the decision variables $w$ contain trajectories of states and controls and the problem data includes terms like control limits or parameters of the system like friction coefficients.

\section{CALIPSO} \label{calipso}
CALIPSO is a differentiable primal-dual augmented Lagrangian interior-point solver for non-convex optimization problems with second-order cone and complementarity constraints. Its standard problem formulation is:
\begin{equation}
	\begin{array}{ll}
		\underset{x}{\mbox{minimize }}  & c(x; \theta) \\
		\mbox{subject to } & g(x; \theta) = 0, \\
		& h(x; \theta) \in \mathcal{K},
	\end{array}
	\label{calipso_problem}
\end{equation}
with decision variables $x \in \mathbf{R}^n$, problem data $\theta \in \mathbf{R}^d$, equality constraints $g : \mathbf{R}^n \times \mathbf{R}^d \rightarrow \mathbf{R}^m$, and constraints $h : \mathbf{R}^n \times \mathbf{R}^d \rightarrow \mathbf{R}^p$ in cone $\mathcal{K} = \mathbf{R}_{++}^q \times Q_{l_1}^{(1)} \times \dots \times Q_{l_j}^{(j)}$ comprising a $q$-dimensional inequality and $j$ second-order cones, each of dimension $l_i$. Internally, problem \eqref{calipso_problem} is reformulated and additional slack variables $r \in \mathbf{R}^m$ and $s \in \mathbf{R}^p$, associated with the equality and cone constraints, respectively, are introduced, and the following modified problem is formed:
\begin{equation}
	\begin{array}{ll}
		\underset{x, r, s}{\mbox{minimize }}  & c(x; \theta) + \lambda^T r + \frac{\rho}{2} r^T r - \kappa \sum \limits_{i = 1}^p \mbox{log}(s^{(i)}) \\
		\mbox{subject to } & g(x; \theta) - r  =0, \\
		& h(x; \theta) - s = 0, \\
		& s \in \mathcal{K},
	\end{array}
	\label{calipso_internal_problem}
\end{equation}
The modified problem's Lagrangian is:
\begin{multline} 
	L(w; \theta, \lambda, \rho, \kappa) = c(x; \theta) + y^T (g(x; \theta) - r) + z^T (h(x; \theta) - s) \\
	+ \lambda^T r + \frac{\rho}{2} r^T r - \kappa \sum \limits_{i = 1}^p \mbox{log}(s^{(i)}). \label{calipso_lagrangian}
\end{multline}
For convenience we denote the concatenation of all of the solver's variables as $w = (x, r, s, y, z, t)$. The KKT system is:
\begin{align} 
	\begin{bmatrix} 
		L_{xx} + \epsilon_p I & 0      & 0 & g_x^T & h_x^T &  0 \\ 
		0      & (\rho + \epsilon_p) I & 0 &    -I &    0 &  0 \\ 
		0      & 0      & \epsilon_p I &    0 &   -I & -I \\ 
		g_x    & -I     & 0 &    -\epsilon_d I &    0 &  0 \\ 
		h_x    & 0      & -I & 0  & -\epsilon_d I & 0 \\
		0 & 0 & P_s & 0 & 0 & P_t - \epsilon_d I
	\end{bmatrix}
	\begin{bmatrix}
		\Delta x \\ 
		\Delta r \\ 
		\Delta s \\ 
		\Delta y \\ 
		\Delta z \\ 
		\Delta t
	\end{bmatrix} \label{calipso_kkt}
	&= 
	- 
	\begin{bmatrix} 
		c_x + g_x^T y + h_x^T z \\ 
		\lambda + \rho r - y \\ 
		-z - t \\ 
		g - r \\ 
		h - s \\
		s \circ t - \kappa \mathbf{e}
	\end{bmatrix}, \notag \\
	J \Delta w &= - R,
\end{align}
where and $P_s, P_t \in \mathbf{S}^p$ are the cone-product Jacobians, and $\epsilon_p, \epsilon_d \in \mathbf{R}_+$ are additional primal and dual regularization terms, respectively.

\subsection{Search direction}
The nominal KKT system \eqref{calipso_kkt} without regularization (i.e., $\epsilon_p, \epsilon_d = 0$) is non-symmetric, potentially with undesirable eigenvalues that will not return a descent direction. The solver modifies this system for faster computation and a more reliable search direction.

\paragraph{Symmetric KKT system.}
To be amenable to fast solvers for symmetric linear systems, the nominal KKT system is reformulated:
\begin{align} 
	\begin{bmatrix} 
		L_{xx} + \epsilon_p I & g_x^T & h_x^T \\ 
		g_x & -\Big(\frac{1}{\rho + \epsilon_p} I \Big) & 0 \\ 
		h_x & 0 & -\Big(\epsilon_d I + ((P_s + \epsilon_p \bar{P}_t)^{-1} \bar{P}_t) \Big)
	\end{bmatrix} 
	\begin{bmatrix} 
		\Delta x \\ 
		\Delta y \\ 
		\Delta z \\
	\end{bmatrix} 
	&=
	-\begin{bmatrix} 
		L_x \\ 
		\bar{L}_y\\ 
		\bar{L}_z \\
	\end{bmatrix}, \label{calipso_search_direction_xyz}
\end{align} 
using Schur complements \cite{boyd2004convex}, where $\bar{L}_y = L_y + \frac{1}{\rho + \epsilon_p} L_r$, $\bar{L}_z = L_z + (P_s + \epsilon_p \bar{P}_t)^{-1}(\bar{P}_t L_s + L_t)$, and $\bar{P}_t = P_t - \epsilon_d I$. The remaining search directions:
\begin{align} 
	\Delta r &= (\Delta y + L_r) / (\rho + \epsilon_p), \label{calipso_search_direction_r}\\
	\Delta s &= (P_s + \epsilon_p \bar{P}_t)^{-1} (\bar{P}_t \Delta z + \bar{P}_t L_s + L_r), \label{calipso_search_direction_s}\\
	\Delta t &= - \Delta z + \epsilon_p \Delta s - L_s, \label{calipso_search_direction_t}
\end{align}
are recovered from the solution \eqref{calipso_search_direction_xyz}. 
Iterative refinement \cite{nocedal2006numerical} is performed to improve the quality of the search directions computed using the symmetric system (\ref{calipso_search_direction_xyz} - \ref{calipso_search_direction_t}).

\paragraph{Inertia correction.}
To ensure a unique descent direction, the left-hand side of the symmetric KKT system \eqref{calipso_search_direction_xyz} is corrected to have an inertia of $n$ positive, $m + p$, negative, and no zero-valued eigenvalues. This is accomplished via adaptive regularization that increases the regularization terms, and subsequently reduces the values when possible to limit unnecessary corrections, using a heuristic developed for Ipopt \cite{wachter2006implementation}.

\subsection{Line search}
Cone variables are initialized being strictly feasible at the start of each solve. A filter line search that is a slight modification of the one used by Ipopt \cite{wachter2006implementation} is performed to ensure an improvement to the solution is achieved at each step of the algorithm and step sizes are initially chosen using a fraction-to-the-boundary rule to ensure that the cone constraints remain strictly satisfied. Additionally, a separate step size is computed for the candidate cone dual variables $t$ to avoid unnecessarily restricting progress of the remaining variables.

After satisfying the cone constraints, a filter line search is performed with the merit function:
\begin{equation}
	\varphi(x, r, s; \theta, \lambda, \rho, \kappa) = c(x; \theta) + \lambda^T r + \frac{\rho}{2} r^T r - \kappa \sum \limits_{i = 1}^p \mbox{log}(s^{(i)}), \label{calipso_merit}
\end{equation}
and violation metric:
\begin{equation} 
	\eta = \|(g(x) - r, h(x) - s) \|_1 / (m + p). \label{slack_constraint_violation}
\end{equation}
The step size is further decremented until either the merit or violation metric is decreased \cite{wachter2005line}. A filter: 
\begin{equation}
\mathcal{F} = \{(\varphi^{(1)}, \eta^{(1)}), \dots, (\varphi^{(p)}, \eta^{(p)})\},
\end{equation}
comprises a set of $p$ previously accepted points. 
A candidate point must satisfy:
\begin{equation} 
  \hat{\varphi} < \varphi^{(i)} \, \lor \, \hat{\eta} < \eta^{(i)}, \quad i = 1, \dots, v,
\end{equation}
for each previous point in the filter. Then, the Armijo condition \cite{nocedal2006numerical}:
\begin{equation} 
  \hat{\varphi} < \varphi + \epsilon_a \alpha ( \varphi_x^T \Delta x + \varphi_r^T \Delta r + \varphi_s^T \Delta s), \label{calipso_armijo}
\end{equation}
with tolerance $\epsilon_a \in \mathbf{R}_+$, must be satisfied in order to accept a candidate point. Finally, the filter is augmented with the candidate point:
\begin{equation} 
  \mathcal{F} \leftarrow \mathcal{F} \cup (\hat{\varphi}, \hat{\eta}).
\end{equation}

\subsection{Cone-product Jacobians} \label{calipso_cone_product_jacobians}
The cone-product Jacobian's $P_s$ and $P_t$ have known structure and decompose by cone. This enables fast matrix products and inverses. For inequalities: 
\begin{equation} 
	P_a(a, b) = \mbox{\textbf{diag}}(b), \quad P_a(a, b)^{-1} = \mbox{\textbf{diag}}(1 / b_1, \dots, 1 / b_q),
\end{equation} 
are diagonal matrices. For second-order cones:
\begin{equation} 
	P_a(a, b) = \begin{bmatrix} b^{(1)} & (b^{(2:l)})^T \\ b^{(2:l)} & b^{(1)} I \end{bmatrix},
\end{equation} 
is an arrowhead matrix and its inverse has complexity linear in the dimension of the matrix \cite{najafi2014efficient}.

\subsection{Fraction-to-the-boundary} \label{calipso_f2b} 
Step sizes $\alpha$ for cone-variable updates are selected to ensure that the fraction-to-the-boundary rule \cite{wachter2006implementation}:
\begin{align}
	s + \alpha \Delta s - (1 - \tau) s \in \mathcal{K}, \label{calipso_fraction_to_boundary}
\end{align}
is satisfied for a cone. The parameter $\tau \in [0, 1]$ helps prevent cone variables from reaching their respective boundaries too quickly and is increased during outer updates. The fraction-to-the-boundary value is increased during each outer update.

\subsection{Iterative refinement}
A drawback to computing search directions using the symmetric system is the potential worsening of the numerical conditioning of the system. As a result, the error:
\begin{equation} 
	e = R + J \Delta w, \label{calipso_search_direction_error}
\end{equation}
will be nonzero. To account for this, we correct the search direction by performing iterative refinement \cite{nocedal2006numerical}. A linear system:
\begin{equation}
	J \Delta e = -e,
\end{equation}
is solved using the error as the residual in order to compute a correction $\Delta e$. This correction is then utilized to update the search direction:
\begin{equation} 
	\Delta w \leftarrow \Delta w + \Delta e. \label{calipso_search_direction_correction}
\end{equation}
This procedure (\ref{calipso_search_direction_error} - \ref{calipso_search_direction_correction}) is repeated until the norm of the error is below a desired tolerance. 

In the case where iterative refinement fails (e.g., exceeding the solver's maximum number of refinement iterations), new candidate points are evaluated using an alternative search direction computed using the regularized non-symmetric system and an LU factorization.

\subsection{Outer updates}
Convergence of a subproblem \eqref{calipso_internal_problem} occurs for fixed values of $\lambda$, $\rho$, and $\kappa$, when the criteria: $\| R \|_{\infty} \leq \gamma_{\kappa} \kappa$,
is met for $\gamma_{\kappa} \in \mathbf{R}_+$. This criteria does not require strict satisfaction of subproblems and decreases the total number of iterations required by the solver. Outer updates on the central-path parameter and the penalty value are subsequently performed:
\begin{equation}
	\kappa \leftarrow  \mbox{max}\Big(\kappa_{\mbox{min}}, \mbox{min}(\psi_{\kappa} \cdot \kappa, \kappa^{\zeta_{\kappa}})\Big), \quad
	\rho \leftarrow \mbox{min}\Big(\rho_{\mbox{max}}, \mbox{max}(\phi_{\rho} \cdot \rho, 1 / \kappa) \Big). \label{calipso_outer_update}
\end{equation}
The updates are clipped to prevent unnecessarily small/large values.

\subsection{Initialization}

Given an initial guess for the primal variables $x_{\mbox{init}}$, the solver's variables are initialized with:
\begin{equation} 
  r = g(x_{\mbox{init}}), \,
  s_{\mbox{ineq}} = \mathbf{1}_q, \,
  s_{\mbox{so}}^{(i)} = (1, 1 / 10 \cdot \mathbf{1}_{l-1}), \,
  y = 0_m, \,
  z = 0_p, \,
  t = s, \label{calipso_initialize}
\end{equation}
where $s_{\mbox{ineq}}$ and $s_{\mbox{so}}$ are inequality and second-order variables, respectively, and are initialized strictly feasible. While more complex schemes may be effective, particularly for specific problems, this simple initialization works well in practice as a default setting.

\subsection{Solution derivatives}
The solution $w^*(\theta)$ returned by CALIPSO is differentiable with respect to its problem data $\theta$. At a solution point, the residual is approximately zero \eqref{implicit_function}, and sensitivities \eqref{solution_sensitivity} are computed using $\partial R / \partial w = J$, which has already been computed and factorized, and:
\begin{equation} 
	\partial R / \partial \theta = (
	L_{x\theta}, 
	0_{m \times d}, 
	0_{p \times d},
	g_{\theta},
	h_{\theta},
	0_{p \times d}).
\end{equation}
Additionally, the sensitivity of the solution with respect to each element of the problem data can be computed in parallel.

\subsection{Implementation}
The CALIPSO solver is summarized in Algorithm \ref{calipso_algorithm}. 

\begin{algorithm}[H]
	\caption{CALIPSO}\label{calipso_algorithm}
	\begin{algorithmic}[H]
		\Procedure{Optimize}{$x, \theta, c, g, h, \mathcal{K}, \gamma$}
		\State \textbf{Parameters}: $\kappa = 1, \rho = 1, \lambda = 0$
		\State \textbf{Initialize}: $r, s, y, z, t$
		\State \textbf{Until} $\| R(w; \theta, y, \infty, 0) \|_{\infty} < \gamma_R$
		\State \indent \textbf{Until} $\|R(w; \theta, \lambda, \rho, \kappa)\|_{\infty} < \gamma_{\kappa} \kappa$ \textbf{do}
		\State \indent \indent $\mathbf{inertia}$ $\mathbf{correction}$: $\epsilon_p$, $\epsilon_d$ \Comment{Eq. (\ref{calipso_kkt})}
		\State \indent \indent $\mathbf{search}$ $\mathbf{direction}$: $\Delta w = (\Delta x, \Delta r, \Delta s, \Delta y, \Delta z, \Delta t)$ \Comment{Eqs. (\ref{calipso_search_direction_xyz} - \ref{calipso_search_direction_t})}
		\State \indent \indent $\mathbf{cone}$ $\mathbf{line}$ $\mathbf{search}$: $\alpha$, $\alpha_t$
		\State \indent \indent $\mathbf{candidate}$: $\hat{x} = x + \alpha \Delta x$, $\hat{r} = r + \alpha \Delta r$, $\hat{s} = s + \alpha \Delta s$
		\State \indent \indent $\mathbf{filter}$: $\hat{\varphi}$, $\hat{\eta}$
		\State \indent \indent $\mathbf{update}$: $w \leftarrow (\hat{x}, \hat{r}, \hat{s}, y + \alpha \Delta y, z + \alpha \Delta z, t + \alpha_t t)$
		\State \indent $\mathbf{outer}$ $\mathbf{update}$: $\lambda, \rho, \kappa$ \Comment{Eq. (\ref{calipso_outer_update}, \ref{augmented_lagrangian_update})}
		\State $\mathbf{Differentiate}$: $\partial w / \partial \theta$ \Comment{Eq. \ref{solution_sensitivity}}
		\State \textbf{Return} $w, \partial w / \partial \theta$
		\EndProcedure
	\end{algorithmic}
\end{algorithm}
An open-source implementation of the solver, \texttt{CALIPSO.jl}, written in Julia, is provided. The solver, and the following examples, are available at: 
\begin{center}
	\url{https://github.com/thowell/CALIPSO.jl}. 
\end{center}

Transcribing problems for CALIPSO requires specifying the objective and constraint functions, and the number of primal variables. Trajectory-optimization problems are automatically formulated into the standard form \eqref{calipso_problem}. Gradients and sparse Jacobians are generated symbolically using the Julia package \texttt{Symbolics.jl}.

\section{Results} \label{results}
We highlight the capabilities of CALIPSO by optimizing a collection of motion-planning problems from manipulation, locomotion, and aerospace domains that require second-order cone and complementarity constraints while transcribing constraints without approximation. Next, we demonstrate the ability to differentiate through the solver and auto-tune policies for non-convex underactuated robotic systems. Additional details about the experimental setups are available in the open-source implementation. A collection of non-convex problems are provided in the Appendix.

\subsection{Contact-implicit trajectory optimization}

CALIPSO is utilized to optimize contact-implicit trajectory optimization problems, see the Appendix for additional details. The contact dynamics \cite{posa2014direct} are directly transcribed without modification. Comparisons are performed with Ipopt using the default MUMPS linear-system solver.

\begin{figure}[H]
    \centering
    \captionsetup[subfigure]{justification=centering}
    \begin{subfigure}{.495\textwidth}
    \centering
      \includegraphics[height=2.5cm]{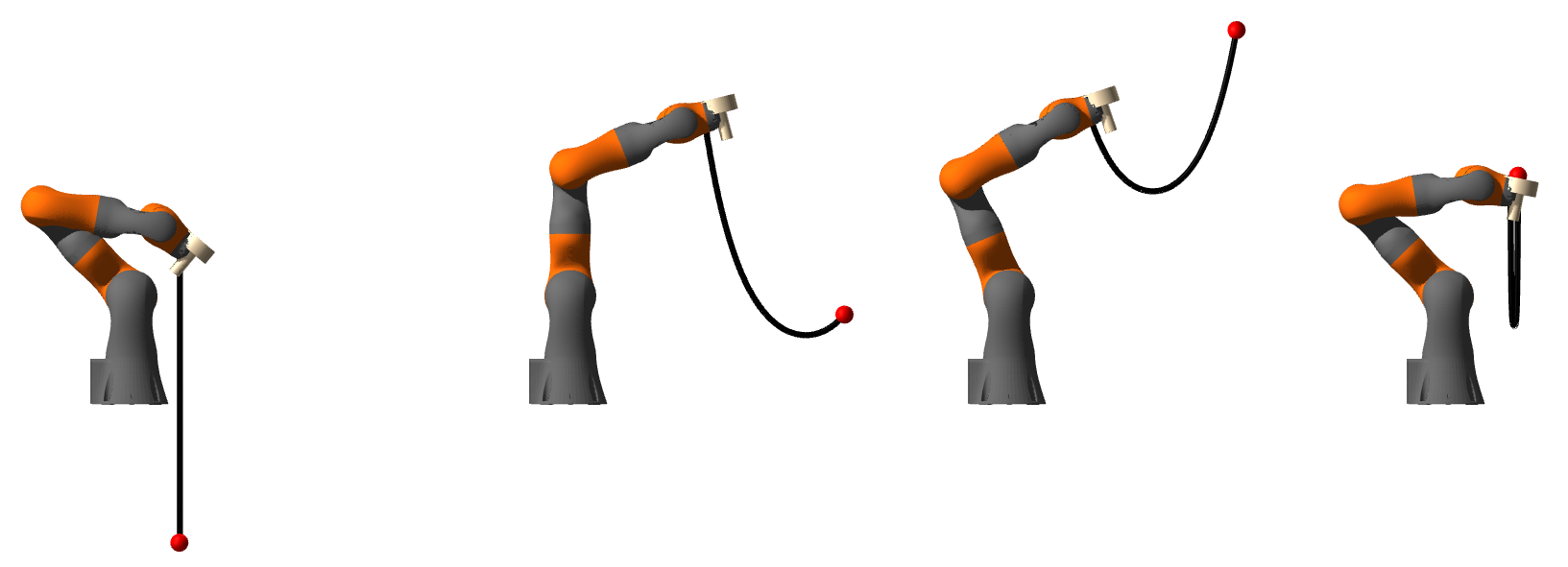}
      \caption{ball-in-cup}
      \label{calipso_ci_ballincup}
    \end{subfigure}%
    \begin{subfigure}{.495\textwidth}
    \centering
      \includegraphics[height=2.5cm]{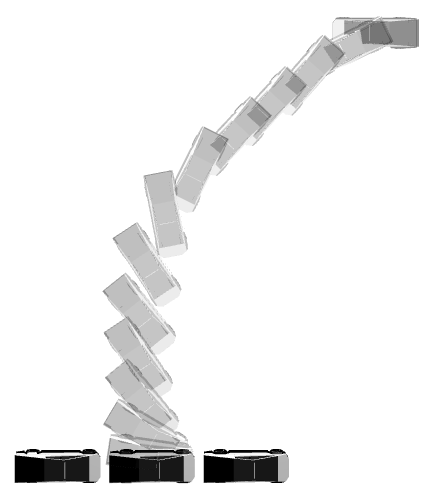}
      \caption{drifting}
      \label{calipso_ci_drifting}
    \end{subfigure}
    \vskip\baselineskip
    \begin{subfigure}{.495\textwidth}
    \centering
      \includegraphics[height=1.25cm]{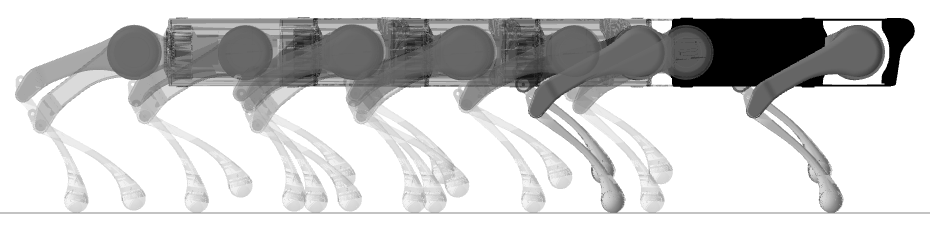}
      \caption{quadruped gait}
      \label{calipso_ci_quadruped_gait}
    \end{subfigure}
    \begin{subfigure}{.495\textwidth}
    \centering
      \includegraphics[height=1.25cm]{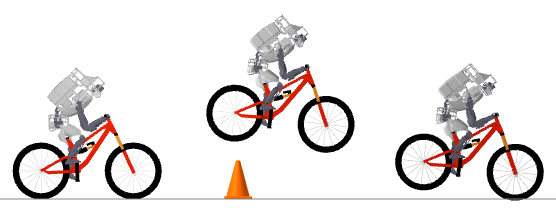}
      \caption{bunny-hop}
      \label{calipso_ci_bunnyhop}
    \end{subfigure}
    \caption[Contact-implicit trajectory optimization examples]{Contact-implicit trajectory optimization examples optimized with CALIPSO. (a) Ball attached to a string is swung into a cup by optimizing end-effector positions and forces. (b) Autonomous car plans a drifting maneuver in order to parallel park. (c) Gait for a quadruped is optimized via a single step and loop constraint. (d) Bicycle robot performs a bunny hop over an obstacle to reach a goal state.}
\end{figure}

\paragraph{Ball-in-cup.} The position and applied forces of a robotic manipulator's end-effector are optimized to swing a ball into a cup (Fig. \ref{calipso_ci_ballincup}). A string between the end-effector and ball is modeled with inequality and complementarity constraints. CALIPSO finds a physically realistic motion plan that is verified with inverse kinematics in simulation, while the Ipopt solution is of poor quality and violates physics by applying nonnegligible force to the ball while the string is slack.

\paragraph{Drifting.}
A parallel-park maneuver is planned that requires an autonomous vehicle to drift (i.e., plan a trajectory with both sticking and sliding contact) into a goal configuration between two parked vehicles (Fig. \ref{calipso_ci_drifting}). The system is modeled as a Dubins car \cite{lavalle2006planning} with Coulomb friction applied to the wheels. Ipopt exhibits extremely poor converge and violates the friction cones to solve this problem, whereas CALIPSO finds a high-quality solution that leverages the nonlinear friction cone to slide into the narrow parking spot.

\paragraph{Quadruped gait.} 
A gait is planned for a planar quadruped by optimizing a single step with a loop constraint (Fig. \ref{calipso_ci_quadruped_gait}). Ipopt struggles to converge, returning a solution with large complementarity violations and a reference that is unusable for online tracking. In contrast, CALIPSO finds a reference trajectory that satisfies the contact dynamics.

\paragraph{Bunny-hop.}
A bicycle robot performs a bunny-hop over an obstacle (Fig. \ref{calipso_ci_bunnyhop}). The rider is modeled as a mass with actuated prismatic joints attached to the bike at each wheel. CALIPSO is able to converge to a trajectory where the bicycle hops over the obstacle by manipulating the rider mass, while Ipopt takes an order of magnitude more iterations to converge.

In summary, Ipopt struggles to return solutions that are useful for robotics applications, whereas CALIPSO reliably returns high-quality solutions while using exact constraint specifications. Numerical results are summarized in Table \ref{calipso_contact_implicit_trajopt_comparison}.

\begin{table}[H]
	\captionsetup[subtable]{justification=centering}
	\centering
	\caption[Numerical comparison of CALIPSO and Ipopt for contact-implicit trajectory optimization]{Comparison between CALIPSO and Ipopt for final objective value, constraint violation, and total iterations on contact-implicit trajectory optimization problems. Cases that failed to converge (i.e., Ipopt falling back to restoration mode) are highlighted in red. Without user-tuned smoothing and problem reformulations, Ipopt performs poorly on these examples and returns solutions that are unusable for robotics applications. In contrast, CALIPSO returns high-quality solutions and does not require approximating constraints.}
	\small
	\begin{subtable}[b]{0.55\linewidth}
		\centering
		\begin{tabular}{c c c c}
			\toprule
			\textbf{Solver} &
			\textbf{Objective} &
			\textbf{Violation} &
			\textbf{Iterations} \\
			\toprule
			Ipopt & \color{red}{68.18} & \color{red}{2.60e{-}2} & \color{red}{205} \\
			CALIPSO & \textbf{11.96} & \textbf{4.86e{-}5} & \textbf{131} \\
			\toprule
		\end{tabular}
		\caption{ball-in-cup}
	 \vspace{1em}
	\end{subtable}
	\begin{subtable}[b]{0.55\linewidth}
		\centering
		\begin{tabular}{c c c c}
			\toprule
			\textbf{Solver} &
			\textbf{Objective} &
			\textbf{Violation} &
			\textbf{Iterations} \\
			\toprule
			Ipopt & \color{red}{8.66} & \color{red}{1.00e{-}1} & \color{red}{194} \\
			CALIPSO & \textbf{0.24} & \textbf{1.38e{-}5} & \textbf{189} \\
			\toprule
		\end{tabular}
		\caption{drifting}
		\vspace{1em}
	\end{subtable}
	\vfill
	\begin{subtable}[b]{0.55\linewidth}
		\centering
		\begin{tabular}{c c c c}
			\toprule
			\textbf{Solver} &
			\textbf{Objective} &
			\textbf{Violation} &
			\textbf{Iterations} \\
			\toprule
			Ipopt & \color{red}{1855.18} & \color{red}{1.13e{-}1} & \color{red}{2000} \\
			CALIPSO & \textbf{574.84} & \textbf{5.32e{-}4} & \textbf{178} \\
			\toprule
		\end{tabular}
		\caption{quadruped gait}
		\vspace{1em}
	\end{subtable}
	\begin{subtable}[b]{0.55\linewidth}
		\centering
		\begin{tabular}{c c c c}
			\toprule
			\textbf{Solver} &
			\textbf{Objective} &
			\textbf{Violation} &
			\textbf{Iterations} \\
			\toprule
			Ipopt & 1503.42 & \textbf{8.96e-8} & 1409 \\
			CALIPSO & \textbf{462.61} & 1.76e-6 & \textbf{101} \\
			\toprule
		\end{tabular}
		\caption{bunny-hop}
	\vspace{1em}
	\end{subtable}
	\label{calipso_contact_implicit_trajopt_comparison}
\end{table}

\subsection{State-triggered constraints}
\begin{figure}[H]
	\centering
	\includegraphics[width=0.5\textwidth]{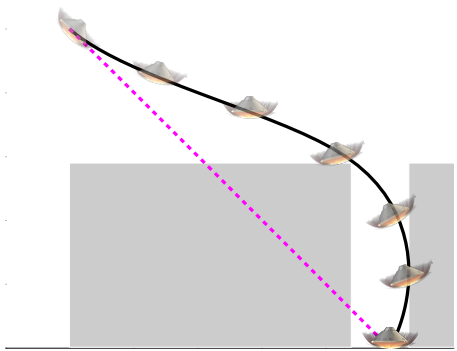}
	\caption{Entry-vehicle soft-landing plan that must avoid elevated regions to the left and right of the landing zone (gray), represented as state-triggered constraints. The unconstrained and constrained solutions are shown in magenta and black respectively.}
	\label{calipso_stc}
\end{figure}

A trigger condition $\Gamma:\mathbf{R}^{n} \rightarrow \mathbf{R}$ encodes the logic: $\Gamma(x) > 0 \implies h(x) \geq 0$, that a constraint is enforced only when the trigger is satisfied. Such state-triggered constraints are utilized within various aerospace applications \cite{szmuk2019real,szmuk2020successive} and commonly utilize a non-smooth formulation (left):
\begin{equation}
\mbox{min}(0, -\Gamma(x)) \cdot h(x) \leq 0
\quad
\rightarrow 
\quad
\begin{aligned}
	\Gamma_+ - \Gamma_- &= \Gamma(x),\\ 
	h_+ - h_- &= h(x), \\
	\Gamma_+ \cdot h_- &= 0, \\
	\Gamma_+, \Gamma_-, h_+, h_- & \geq 0, 
\end{aligned}
\end{equation}
that linearizes poorly and can violate LICQ. In this work, we employ an equivalent complementarity formulation (right) to land an entry-vehicle in an environment with keep-out zones adjacent to the landing site. The resulting solution (Fig. \ref{calipso_stc}) demonstrates CALIPSO's ability to find solutions that satisfy complementarity constraints. 

Ipopt returns a solution that violates the keep-out zone for the much of the trajectory. 

\subsection{Model predictive control auto-tuning}
    
CALIPSO is utilized offline to plan reference trajectories and online as the tracking controller in a model predictive control (MPC) policy \cite{richalet1978model}. The policy aims to track the reference, given an updated state estimate from a simulation, by re-planning over a reduced horizon to compute a new (feedback) control that is then applied to the system.

The policy's cost weights are treated as problem data to be optimized and the solution computed by the policy (i.e., the re-optimized control inputs) are differentiated with respect to these parameters \eqref{solution_sensitivity} in order to compute gradients that are utilized to automatically tune the policy. The metric for tuning the policy consists of quadratic costs on tracking the reference. Gradient descent with a line search is used to update the cost weights by differentiating through rollouts of the policy. 

The policy weights are initialized with all ones and we compare the performance of the auto-tuned policy after 10 gradient steps with open-loop and untuned policies on swing-up tasks for cart-pole and acrobot systems. For both systems, the auto-tuned policy outperforms the baselines; results are summarized in Table \ref{calipso_mpc_autotune}.

\begin{table}[H] 
	\centering
	\caption[Numerical results for predictive control auto-tuning]{Comparison of tracking error between open-loop and both untuned and auto-tuned model predictive control policies in simulation. By differentiating through CALIPSO, gradient-based optimization is able to rapidly improve controller performance without requiring designer input.}
	\begin{tabular}{c c c c}
		\toprule
		&
		\textbf{Open-Loop} &
		\textbf{MPC (untuned)} &
		\textbf{MPC (tuned)} \\
		\toprule
		cart-pole & 5.11e4 & 15.06 & \textbf{0.79}\\
		acrobot & 1.38e4 & 439.26 & \textbf{0.04}\\
		\toprule
	\end{tabular}
	\label{calipso_mpc_autotune}
\end{table}

\section{Conclusion} \label{conclusion}
We now discuss the limitations of the work and directions for future research.

\subsection{Limitations} 
In the non-convex setting it is generally not possible to guarantee that the optimizer will find a globally optimal solution. Further, while prior work analyzes the convergence properties for line-search filter methods \cite{wachter2005line}, we leave this analysis for CALIPSO to future work. 

Despite the numerical improvements of the solver, in many cases we still find that contact-implicit trajectory optimization problems are difficult to optimize. In practice we find that good initialization is crucial, but results that generate qualitatively different contact sequences compared to the initialization are rare. 

\subsection{Future Work} 
Future work will add support for additional cones, which naturally fit within CALIPSO's interior-point framework, including semidefinite cones that are of interest in many control applications \cite{palan2020fitting}---particularly settings with nonlinear dynamics. Additionally, extending the Julia implementation to C/C++ will potentially enable real-time performance of MPC policies onboard robots with limited computing hardware. The solver has potential to support state-triggered constraints online in safety-critical applications or be utilized in feedback loops for contact-implicit model predictive control \cite{lecleach2021fast} with systems that make and break contact with their environments.

In conclusion, we have presented a new solver for trajectory optimization problems with second-order cone and complementarity constraints: CALIPSO. The solver prioritizes reliability and numerical robustness, and offers specialized constraint support that enables planning for challenging tasks arising in manipulation, locomotion, and aerospace applications while enabling the designer to exactly transcribe constraints without requiring problem reformulations. Additionally, the solver is differentiable with respect to its problem data, allowing it to be called by efficient, gradient-based, upper-level optimization routines for applications like policy auto-tuning. To the best of our knowledge, no existing solver offers this collection of unique features. 

\section*{Acknowledgments}
The authors would like to thank Shane Barratt for helpful discussions related to cone programming. 
This work was supported by Frontier Robotics, Innovative Research Excellence, Honda R\&D Co., Ltd.
and an Early Career Faculty Award from NASA’s Space Technology Research Grants Program (Grant Number 80NSSC21K1329).

\vspace{0.5cm}
\begingroup
\let\clearpage\relax
{\footnotesize
\renewcommand\bibpreamble{\vspace{-2.5\baselineskip}} 
\setlength{\bibsep}{0pt} 
\renewcommand{\bibname}{\large References}
\bibliography{main}}
\endgroup

\begingroup
\let\clearpage\relax
\begin{appendices}
\newpage
\chapter{Non-Convex Optimization Problems} \label{calipso_nonconvex_benchmarks}
Three small, non-convex problems are optimized with CALIPSO. 

\paragraph{W\"{a}chter problem.}
Motivating the development of a number of Ipopt's key algorithms, the following problem:
\begin{equation}
	\begin{array}{ll}
		\underset{x_1, x_2, x_3}{\mbox{minimize }}  & x_1\\
		\mbox{subject to } & x_1^2 - x_2 - 1 = 0, \\
		& x_1 - x_3 - \frac{1}{2} = 0, \\
		& x_2, x_3 \geq 0,
	\end{array}
	\label{wachter_problem}
\end{equation}
when initialized with a point, $x_1 < 0$, $x_2 > 0$, $x_3 > 0$, causes many infeasible-start interior-point methods to fail \cite{hinder2018one}. Given the point $x_{\mbox{init}} = (-2, 3, 1)$, CALIPSO finds the optimal solution $x^* = (1, 0, \frac{1}{2})$ in 17 iterations.

\paragraph{Maratos problem.}
The following problem: 
\begin{equation}
	\begin{array}{ll}
		\underset{x_1, x_2}{\mbox{minimize }}  & 2 (x_1^2 + x_2^2 - 1) - x_1,\\
		\mbox{subject to } & x_1^2 + x_2^2 - 1 = 0, \\
	\end{array}
	\label{maratos_problem}
\end{equation}
highlights the Maratos effect \cite{nocedal2006numerical}, which often requires a solver to perform second-order corrections. CALIPSO's use of an augmented Lagrangian for the equality constraints allows for the omission of second-order corrections, allowing convergence on this problem from a starting point $x_{\mbox{init}} = (2, 1)$ to the optimal solution $x^* = (1, 0)$ in 6 iterations.

\paragraph{Complementarity problem.}
Challenging complementarity-constrained problems such as:
\begin{equation}
	\begin{array}{ll}
		\underset{x}{\mbox{minimize }} & (x_1 - 5)^2 + (2x_2 + 1)^2 \\
		\mbox{subject to } & 2 (x_2 -1) - \frac{3}{2} x_2 + x_3 - \frac{1}{2} x_4 + x_5 = 0, \\
		& 3 x_1 - x_2 - x_6 - 3 = 0, \\
		& -x_1 + \frac{1}{2} x_2 - x_7 + 4 = 0, \\ 
		& -x_1 - x_2 - x_8 + 7 = 0, \\ 
		& x_3 \cdot x_6 = 0, \\
		& x_4 \cdot x_7 = 0, \\ 
		& x_5 \cdot x_8 = 0, \\
		& x \geq 0,
	\end{array}
	\label{knitro_problem}
\end{equation}
often require manual reformulations in order to be solved by general-purpose optimizers. With CALIPSO, we directly specify the problem, initialize the solver with $x_{\mbox{init}} = 0$, and find the optimal solution $x^* = (1, 0, 2, 0, 0, 0, 3, 6)$ in 12 iterations.

\chapter{Contact-Implicit Trajectory Optimization}
Direct trajectory optimization plans trajectories for systems that make and break contact with their environment without requiring hybrid dynamics \cite{westervelt2003hybrid} or explicitly enumerating all of the possible sequences of contact configurations by utilizing complementarity-based contact dynamics formulations \cite{stewart1996implicit} as explicit constraints \cite{posa2014direct}. This enables the optimizer to potentially generate motion plans without pre-specified contact plans using task-level specifications via an objective.
 
\paragraph{Problem.} The contact-implicit trajectory optimization problem:
\begin{align}
	\underset{\substack{x_{1:T}, u_{1:T-1}, \\ \gamma_{1:T-1}, \beta_{1:T-1}, \eta_{1:T-1}}}{\mbox{minimize }} & \quad c_T(x_T) + \sum \limits_{t = 1}^{T-1} c_t(x_t, u_t) \label{intro_ci_trajopt}\\
	\mbox{subject to} & \quad S(x_t, x_{t+1}) = B(q_{t}) u + J(q_{t})^T \lambda_t, \quad t = 1,\dots,T-1, \notag \\
	& \quad q_{t+1} = q_t + h v_{t+1}, \quad \quad \quad \quad \quad \quad \quad \,\, \, t = 1,\dots,T-1, \notag \\
	& \quad \gamma_t \cdot \phi(q_{t+1}) = 0, \quad \quad \quad \quad \quad \quad \quad \quad \, \, \, t = 1,\dots,T-1, \notag \\
	& \quad \gamma_t,\phi(q_{t+1}) \geq 0, \quad \quad \quad \quad \quad \quad \quad \quad \mskip0.5\thinmuskip \, \, \, \, t = 1, \dots, T-1, \notag \\
    & \quad D(x_{t+1}) v_{t+1} - \eta_t^{(2:3)} = 0, \quad \quad \quad \quad \, \, t = 1, \dots, T-1, \notag \\
	& \quad \beta_t^{(1)} - \mu \gamma_t = 0, \quad \quad \quad \quad \quad \quad \quad \quad \, \, \, \, \, t = 1, \dots, T-1, \notag\\
	& \quad \beta_t \circ \eta_t = 0, \quad \quad \quad \quad \quad \quad \quad \quad \quad \, \, \, \, \, \, \, \, \, t = 1, \dots, T-1, \notag \\
	& \quad \|\beta_t^{(2:3)}\|_2 \leq \beta_t^{(1)}, \, \|\eta_t^{(2:3)}\|_2 \leq \eta_t^{(1)}, \quad t = 1, \dots, T-1, \notag \\
	& \quad (x_1 \, \mbox{given}) \notag,
\end{align}
with states $x = (q, v) \in \mathbf{R}^{n_q + n_v}$, controls $u \in \mathbf{R}^{n_u}$, integrated smooth dynamics $S : \mathbf{R}^{n_q \times n_v} \times \mathbf{R}^{n_q \times n_v} \rightarrow \mathbf{R}^{n_v}$, input Jacobian $B \in \mathbf{R}^{n_v \times n_u}$, contact Jacobian $J \in \mathbf{R}^{3 \times n_v}$, time step $h \in \mathbf{R}_{++}$, primal and dual friction variables $\beta, \eta \in \mathbf{R}^3$, contact forces $\lambda = (\beta^{(2:3)}, \gamma) \in \mathbf{R}^2 \times \mathbf{R}$, signed distance function $\phi : \mathbf{R}^{n_q} \rightarrow \mathbf{R}$, contact tangent-space projection matrix $D \in \mathbf{R}^{2 \times n_v}$, friction coefficient $\mu \in \mathbf{R}_{+}$, and where $\circ$ is a second-order cone product \eqref{soc_product}. This formulation aims to minimize an objective with stage $c_t : \mathbf{R}^{n_q \times n_v} \times \mathbf{R}^{n_u} \rightarrow \mathbf{R}$ and terminal costs $c_T : \mathbf{R}^{n_q \times n_v} \rightarrow \mathbf{R}$, and represents the dynamics by explicitly encoding the contact dynamics as constraints. This single contact formulation generalizes to multiple contacts.

Because of CALIPSO's ability to handle second-order cone and complementarity constraints, we can directly use the optimality conditions for the Maximum Dissipation Principle with a nonlinear friction cone as constraints to encode friction behavior. For more details on these friction constraints see: \cite{howell2022dojo}.

\paragraph{Complementarity reformulation.} To work well in practice with general-purpose off-the-shelf solvers for non-convex problems, the complementarity constraints are reformulated using an exact $\ell_1\mbox{-norm}$ penalty \cite{manchester2020variational}:
\begin{equation}
	\begin{array}{ll}
		\underset{}{\mbox{find }}  & a, b\\
		\mbox{subject to } & a \circ b = 0, \\ 
		& a, b \geq 0
	\end{array}
	\quad
	\rightarrow 
	\quad
	\begin{array}{ll}
		\underset{a, b, s}{\mbox{minimize }}  & \rho s\\
		\mbox{subject to } & s \mathbf{1} - a \circ b \geq 0, \\
		& a, b, s \geq 0
	\end{array} \label{intro_complementarity_reformulation}
\end{equation}
This formulation relaxes the complementarity constraints and empirically results in superior convergence properties. As $\rho \rightarrow \infty$, we have $s \rightarrow 0$, and the original formulation is recovered. Despite the reformulation's practical performance, it requires additional decision variables and careful selection of the initial penalty parameter, $\rho \in \mathbf{R}_+$.

\end{appendices}
\endgroup

\end{document}